\documentclass[conference, 10pt]{IEEEtran} %
\IEEEoverridecommandlockouts
\usepackage{cite}
\usepackage{authblk}
\usepackage{amsmath,amssymb,amsfonts}
\usepackage{algorithmic}
\usepackage{graphicx}

\usepackage{subcaption}
\usepackage{footnote}
\usepackage{float}
\usepackage{cuted}
\usepackage{comment}
\usepackage[para,flushleft]{threeparttable}
\usepackage{adjustbox}
\usepackage{nicematrix}
\usepackage[shortlabels,inline]{enumitem}
\usepackage{textcomp}
\usepackage{xcolor}
\usepackage{color, colortbl}
\usepackage{bm}

\def\BibTeX{{\rm B\kern-.05em{\sc i\kern-.025em b}\kern-.08em
    T\kern-.1667em\lower.7ex\hbox{E}\kern-.125emX}}
\begin{document}

\makeatletter
\def\bbordermatrix#1{\begingroup \m@th
  \@tempdima 4.75\p@
  \setbox\z@\vbox{%
    \def\cr{\crcr\noalign{\kern2\p@\global\let\cr\endline}}%
    \ialign{$##$\hfil\kern2\p@\kern\@tempdima&\thinspace\hfil$##$\hfil
      &&\quad\hfil$##$\hfil\crcr
      \omit\strut\hfil\crcr\noalign{\kern-\baselineskip}%
      #1\crcr\omit\strut\cr}}%
  \setbox\tw@\vbox{\unvcopy\z@\global\setbox\@ne\lastbox}%
  \setbox\tw@\hbox{\unhbox\@ne\unskip\global\setbox\@ne\lastbox}%
  \setbox\tw@\hbox{$\kern\wd\@ne\kern-\@tempdima\left[\kern-\wd\@ne
    \global\setbox\@ne\vbox{\box\@ne\kern2\p@}%
    \vcenter{\kern-\ht\@ne\unvbox\z@\kern-\baselineskip}\,\right]$}%
  \null\;\vbox{\kern\ht\@ne\box\tw@}\endgroup}
\makeatother

\title{FMM-X3D: FPGA-based modeling and mapping of X3D for Human Action Recognition}

\author[1, 2]{Petros Toupas}
\author[1]{Christos-Savvas Bouganis}
\author[2]{Dimitrios Tzovaras}
\affil[1]{Dept. of Electrical and Electronic Engineering, Imperial College London, UK}
\affil[ ]{\textit{\{p.toupas21,christos-savvas.bouganis\}@imperial.ac.uk}}
\affil[2]{Information Technologies Institute, Centre for Research and Technology Hellas, Greece}
\affil[ ]{\textit{\{ptoupas,Dimitrios.Tzovaras\}@iti.gr}}

\maketitle

\begin{abstract}
3D Convolutional Neural Networks are gaining increasing attention from researchers and practitioners and have found applications in many domains, such as surveillance systems, autonomous vehicles, human monitoring systems, and video retrieval. However, their widespread adoption is hindered by their high computational and memory requirements, especially when resource-constrained systems are targeted. This paper addresses the problem of mapping X3D, a state-of-the-art model in Human Action Recognition that achieves accuracy of 95.5\% in the UCF101 benchmark, onto any FPGA device. The proposed toolflow generates an optimised stream-based hardware system, taking into account the available resources and off-chip memory characteristics of the FPGA device. The generated designs push further the current performance-accuracy pareto front, and enable for the first time the targeting of such complex model architectures for the Human Action Recognition task.
\end{abstract}

\section{Introduction}

In recent years, two-dimensional CNNs have excelled at image-related tasks. The growing focus and number of applications arising from video-related tasks, such as video surveillance, autonomous driving, and elderly/patient monitoring, have necessitated the development of algorithms that include and account for the temporal domain. Three-dimensional CNNs are one of the most frequently applied techniques for dealing with video and volumetric data. With the addition of an extra dimension, such as time or depth, over 2D CNNs, 3D CNNs improve their learning capacity by extracting features relevant to the newly added dimension of the input as well.

\begin{figure}[]
    \caption{Kinetics-400 pareto over the years}
    \centering
    \includegraphics[width=1.0\linewidth,keepaspectratio]{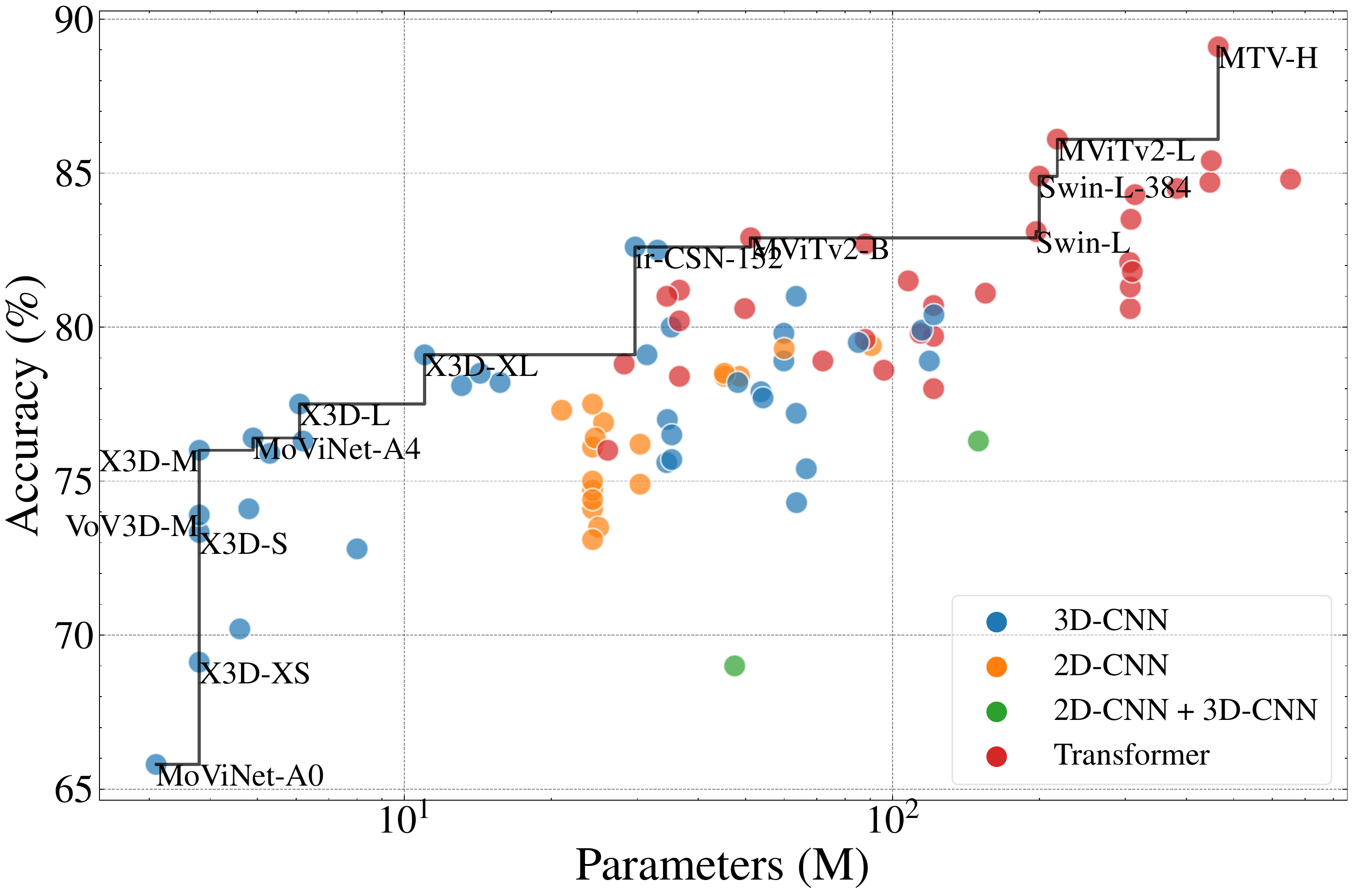}
    \label{fig:pareto_over_years}
\end{figure}

Especially in the task of human action recognition, 3D CNNs have demonstrated exceptional performance. The use of 3D CNNs enables the interpretation of human motion across the frames of a video, enabling the detection of a variety of human actions without the need for dedicated time-domain techniques (e.g., LSTMs). As can be seen in Figure \ref{fig:pareto_over_years}, 3D CNNs dominate the pareto front in one of the most widely used HAR benchmarks, Kinetics-400 \cite{Kay2017TheDataset}, while the recent emergence of vision transformers has also begun to drive some designs to the pareto front; however, such networks require orders of magnitude more GFLOPs to operate.

While 3D CNNs are capable of capturing time- or depth-related features, the additional dimension in the input frequently results in greater workloads and computational and memory requirements as compared to 2D CNNs. Numerous hardware devices, including GPUs, FPGAs, and ASICs, have been used to mitigate the 3D CNNs' high processing requirements and provide high performing systems. FPGAs are an attractive candidate as an acceleration platform since they offer greater flexibility than ASICs and are more energy efficient than GPUs. Furthermore, the rapid evolution of 3D CNN model designs necessitates the need for a platform that can quickly adapt to any new model requirements.

Previously, FPGA-based acceleration of 3D CNN models focused on C3D and R(2+1)D, whose performance falls short of the pareto front in recent benchmarks such as the Kinetics. In contrast, this work proposes an FPGA architecture for X3D\cite{Feichtenhofer2020}, a 3D CNN model that lies in the pareto front of accuracy vs. number of parameters, as shown in Figure \ref{fig:pareto_over_years}. Using an automated optimization approach and by applying transformations to the SDF graph, the design space of X3D is explored based on a streaming architecture and the Synchronous Data Flow (SDF)\cite{Lee1987} computation model, taking into account the target FPGA platform's requirements.

\begin{figure}[]
    \captionsetup{justification=centering,margin=1.25cm}
    \centering
    \includegraphics[width=0.85\linewidth,keepaspectratio]{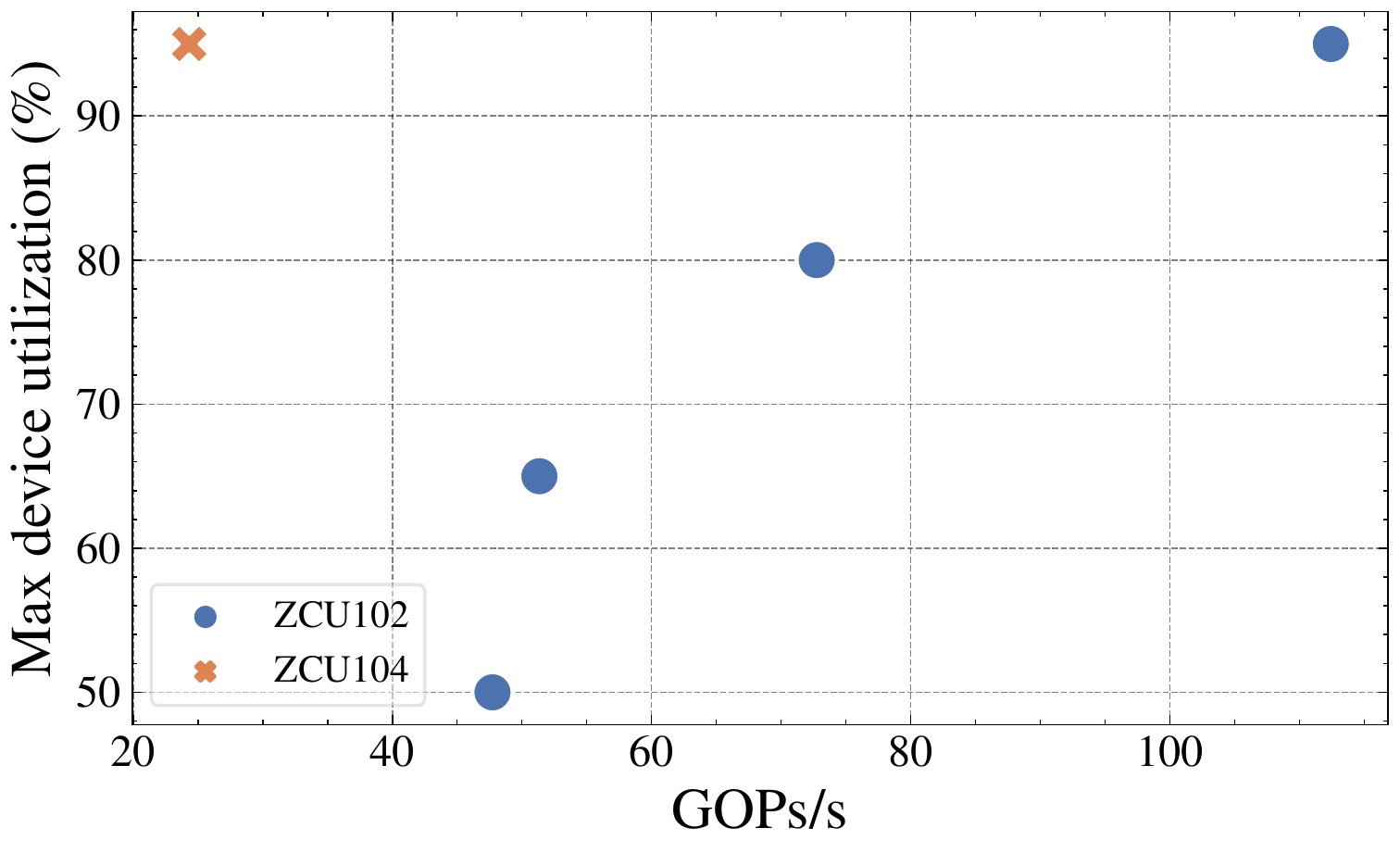}
    \caption{Model performance over different FPGA devices and resource constraints}
    \label{fig:performance_devices}
\end{figure}

The target FPGA platform or resource constraints may change depending on the application or system requirements. As can be seen in Figure \ref{fig:performance_devices}, the proposed hardware design may adapt to different devices as well as varied restrictions regarding the FPGA utilisation percentage, resulting in pareto front solutions in each circumstance. The key contributions of this paper are the following:
\begin{itemize}
    \item The extension of the SDF graph model that is employed for capturing the performance requirements in CNN mapping to streaming architectures to explicitly address irregular blocks with branching, which are widely used in modern CNN models.  
    \item The automation and optimisation of the mapping of the X3D model, a state of the art model for Human Action Recognition, to different FPGA devices, taking into account the available FPGA resources and memory bandwidth characteristics.
    \item Two streaming-centric optimizations to improve the X3D network's dataflow characteristics. The use of prior batch statistics for GAP layers, and the quantization of the model weights and feature maps to 16-bit fixed-point arithmetic.
\end{itemize}

\section{Background}\label{background}

\subsection{fpgaConvNet}

This work is inspired by the fpgaConvNet \cite{Venieris2019}, a framework that automatically maps 2D CNN models to FPGA platforms. The framework accepts a high-level description of a 2D CNN model and the FPGA platform's characteristics. The structure and parameters of the network are captured in a Directed Acyclic Graph (DAG), which is translated into an SDF graph. Synchronous Data-Flow (SDF) \cite{Lee1987} is a computation model for both parallel and sequential systems developed by Lee et. al. in 1987. The dataflow model represents the system's behaviour as a graph, named SDFG. The SDFG nodes represent the operations, while the arcs represent the data streams that connect them. SDF's key idea is that every node fires whenever data is available at its input arcs, resulting in a data-driven execution paradigm. The SDFG is represented as a topology matrix $\Gamma$ and is subjected to a series of transformations that vary the folding factors and reconfiguration points, therefore broadening the design space (a more detailed explanation of how $\Gamma$ matrix is constructed is given in Section \ref{sdfg_representation}). After that, an optimization technique is employed to determine the optimal folding factors and reconfiguration points for optimising the throughput, that result in the final design choice. The tool concludes by providing a synthesizable Vivado HLS hardware design.

\subsection{Related Work}

Although 3D CNNs have been around for a while, there have only been a few papers aimed at their acceleration on FPGAs. The majority of these works focus on relatively old 3D CNNs, such as the C3D \cite{Ji2013} model, whose performance falls short of state-of-the-art models, due to simple underlying CNN model. Fan et al. introduced a series of works on 3D CNN acceleration for HAR on FPGA systems, \cite{Fan2017, Fan2018, Fan2019}. In their initial work \cite{Fan2017}, they proposed the F-C3D hardware architecture for the acceleration of C3D \cite{Ji2013}, which is capable of supporting multiple 3D convolutional layers and design strategies for overcoming the challenges associated with 3D CNNs, such as increased computation and memory requirements, while also allowing their design to be ported to other FPGA devices. In their subsequent work \cite{Fan2018}, they proposed an analytical model and a tool for optimising the hardware architecture based on the device specification and accuracy requirements, as well as the use of block floating point (BFP) arithmetic precision to minimise accuracy loss and the need for retraining the model. Their design evaluation was conducted on C3D model as well. In their most recent publication in this series, \cite{Fan2019}, they proposed E3DNet, an efficient 3D CNN based on their proposed 3D-1 bottleneck building block. Their hardware implementation of E3DNet, called F-E3D, is capable of real-time performance at the execution time of 35.3 ms per clip\footnote{A clip is defined as a stacked sequence of frames that are meant to be the input of the 3D CNN}, while achieving 85.1\%\footnote{H. Duan et. al. \cite{Duan2020} currently holds the SoA results on UCF101 achieving 98.6\% accuracy} in terms of accuracy on the UCF101 benchmark.

Liu et al. \cite{Liu2019AFpgas} proposed a unified hardware architecture for 2D and 3D CNN acceleration based on their observation that the computing patterns of 2D and 3D CNNs are similar. Their aim was to convert the CNN convolutions to matrix multiplication operations, paying close attention to memory optimizations in order to overcome the difficulties of feature map replications. Additionally, they employed an analytical model to configure the accelerators for optimal resource use. They have targeted and evaluated their design on C3D model. Shen et al. \cite{Shen2018} followed a similar approach, developing a unified template-based architecture based on the Winograd \cite{Winograd1980ArithmeticComputations} algorithm capable of handling both 2D and 3D CNNs. Additionally, they developed an analytical technique for efficiently exploring the design space for mapping 2D and 3D CNNs on FPGA accelerators. The authors have targeted the C3D model for the evaluation of their proposed design. Sun et al. \cite{Sun20203DPruning} used a blockwise pruning approach to apply weight pruning to two distinct 3D CNN architectures, namely C3D and R(2+1)\cite{Tran2018}. Their hardware design, which is based on the Alternating Direction Method of Multipliers (ADMM), together with the suggested pruning approach, enables the acceleration of 3D CNNs with low accuracy loss when compared to their unpruned form. Recently, Toupas et al. \cite{ToupasHARFLOW3D:Devices} introduced a toolflow that automates the mapping and optimization of 3D CNN models on FPGA devices achieving promising results on latency-oriented applications.

Until recently, the majority of research has been focused on the C3D \cite{Ji2013} model for HAR, which was introduced in 2013. The model's architecture is rather simplistic, consisting of only six consecutive layers, and it performs poorly in terms of accuracy when compared to the modern SoA models in HAR ($85.2$\% in UCF101 compared to $98.6$\% which is the current SoA). In terms of design complexity, it is comparable to the LeNet or AlexNet in the three-dimensional space. Because the methodologies described above are primarily tailored to the design of the target model, they cannot be extended or evaluated in the more complex architectures of today's state-of-the-art HAR models. As a result, it is unclear how well they would perform and how they could map the most recent SoA models. This study focuses on X3D \cite{Feichtenhofer2020}, which has over 200 layers and deviates from the sequential model of early networks by introducing branching.

\subsection{X3D Model Family}

C. Feichtenhofer proposed the X3D family\cite{Feichtenhofer2020} of efficient 3D CNN models in 2020. The model designs in the family range from X-Small (X3D-XS) to XX-Large (X3D-XXL), depending on the computational complexity, number of layers, input volume size, and number of parameters. Indicative, the computational complexity of the X3D-XS is 0.6 GFLOPs, whereas the computational complexity of the X3D-XXL is 143.5 GFLOPs. The number of model parameters ranges from 3.76 M to 20.3 M. Until 2021, the X3D series had the best performance in many HAR benchmarks, including the Kinetics-400 (80.4\%), Kinetics-600 (81.9\%), and Charades (47.1\%). This paper focuses on the X3D-M, which has a computational complexity of 6.2 GFLOPs, 3.76 M parameters, and receives as input a sequence of 16 RGB frames with spatial dimensions of $256\times256$, attaining an accuracy of 76.0\% in Kinetics-400 and 78.8\% in Kinetics-600. X3D-M has a total of 244 distinct layers, which include depth-wise and point-wise 3D convolutional layers, various activation functions such as relu, sigmoid, and swish, 3D global average pooling layers, and fully connected layers.

\section{Hardware-Level Interpretation of X3D}\label{modelling}

This section discusses the hardware-level interpretation and modelling of the X3D model. The proposed framework takes as input the high-level description file of the X3D model, and captures the parameters of each layer in a DAG along with the connections between the layers. Each of the network's supported layer is mapped to a hardware building block equivalent that implements its functionality. Subsequently the framework generates the SDFG for the network, which is created by mapping the DAG nodes to their hardware equivalents and adding them as nodes in the SDFG, as well as by adding the connections between the DAG nodes as arcs in the SDFG. Finally, the estimated performance of a particular configuration of the SDFG nodes of the network is calculated using the SDF computation model. The following sections provide detailed descriptions for each stage of the proposed framework's workflow.

\subsection{X3D layers as DAG nodes}

Three main parts comprise the description of a neural network model supplied by high-level frameworks such as pytorch and onnx. First, the layers and their connections that define the model's structure and flow. Second, each layer's special attributes and configuration, and finally the actual values of the learnable parameters associated with their layers (if any). A dedicated model parser is provided, that parses the above descriptions to build a DAG containing all of the relevant information about the target neural network. The DAG structure is faithful to the original, retaining just the essential information from the layers' specific attributes and configuration. Additionally, the parser stores the model parameters/weights for future use during inference.

All the layers of the X3D model comprise of the $\mathbf{Sp_i}$, and $\mathbf{Sp_o}$ parameters describing the input and output shapes. In addition to these, each layer has layer-specific parameters, as shown in the configurations below:
\begin{itemize}
    \item \textbf{3D Convolutional Layer} 
        There are four different types of convolutional layers that X3D model incorporates: 3D convolutions with kernel omitting \begin{enumerate*}[(a)]
            \item the temporal dimension $1 \times K_h \times K_w$,
            \item the spatial dimensions $K_d \times 1 \times 1$,
            \item 3D depth-wise convolutions,
            \item 3D point-wise convolutions.
        \end{enumerate*}
        Even though the are different types of convolutions, the discrimination between them can be achieved by looking at the layer's configuration that is stored in the DAG nodes, $\mathbf{<Sp_i, Sp_o, Ks, Sr, Pad, Gp>}$:
        \begin{itemize}[label=$\circ$,leftmargin=+.2in]
            \item \textit{Ks} is a 3-value vector $[K_d, K_h, K_w]$ specifying the depth, height and width of the 3D convolution window (also known as kernel).
            \item \textit{Sr} is a 3-value vector specifying the strides of the convolution along each dimension.
            \item \textit{Pad} is a 3-value vector denoting the amount of padding applied to each dimension of the input
            \item \textit{Gp} a value that specifies the number of groups in which the input is split along the channel axis
        \end{itemize}
    \item \textbf{3D Activation Layers} 
        The activation functions used in X3D are the following: \begin{enumerate*}[(a)]
            \item ReLu activation,
            \item Sigmoid activation,
            \item Swish activation which is expressed as $ y = x * sigmoid(x) $.
        \end{enumerate*}. The layer's configuration is as follows, $\mathbf{<Sp_i, Sp_o, T>}$:
        \begin{itemize}[label=$\circ$,leftmargin=+.2in]
            \item \textit{T} denotes the type of activation function.
        \end{itemize}
    \item \textbf{3D Element-wise Layers} 
        In X3D, element-wise operations refer to the layers that combine data from several branches via addition or multiplication. In comparison to the other layer types, these layers combine several inputs into a single output. The shapes of the layer's inputs may or may not be identical, resulting in different functionality of the layer (normal vs broadcasting). The layer's configuration is as follows, $\mathbf{<Sp_{i1}, ..., Sp_{iN}, Sp_o, T, M>}$:
        \begin{itemize}[label=$\circ$,leftmargin=+.2in]
            \item \textit{T} denotes the type of element-wise operation, e.g. addition, multiplication, division, and so on.
            \item \textit{M} denotes the mode of element-wise operation, normal or broadcasting.
        \end{itemize}
    \item \textbf{3D Global Average Pooling Layer} 
    While the standard pooling operation samples patches of the input feature map to decrease its size, this extreme case samples the whole feature map into a single value, creating an output vector with the same shape as the channels, with the configuration being like, $\mathbf{<Sp_i, Sp_o, T>}$.
\end{itemize}

\subsection{SDFG representation with branch support}\label{sdfg_representation}

To take advantage of the SDF model's capabilities, DAG nodes are mapped into their associated hardware building blocks, which implement the functionality of the respective network layer in the underlying hardware. Using SDF theory, the SDFG may be represented as a topology matrix $\Gamma$. The nodes are represented by the columns of this matrix, while the arcs that link the nodes are represented by the rows. The data consumption/production rates for each node in each arc can be inferred by looking the element at $(node, arc)$ position in the $\Gamma$ matrix. Positive values, by convention, drive data production, whereas negative ones drive data consumption. The element $\Gamma(n,a)=-1$, for example, indicates that node n consumes data at arc a at a rate of one.

The $\Gamma$ matrix is decomposed into several matrices, allowing a more in-depth examination of each one separately and more fine control overall. The initial decomposition of the $\Gamma$ matrix yielded three distinct matrices, which are as follows:
\begin{enumerate}[i)]
    \item The stream matrix $\textbf{S}$. This matrix element stores the number of incoming and outgoing parallel streams that arrive to each node's input and output.
    \item The rate matrix $\textbf{R}$. The rate matrix elements include the normalised data production and consumption rates of each node at each arc (defined as the number of elements produced/consumed per cycle). The values in this matrix range from 0 to 1.
    \item The data matrix $\textbf{C}$. The width of each individual stream from the $\textit{S}$ matrix is stored in this matrix elements. Since all of the streams are assumed to have the same bit width of 16, the above matrix will not be used in this study.
\end{enumerate}

The $\Gamma$ matrix decomposition can be derived from the following Equation: $\mathbf{\Gamma = S \times R}$.

\small
\setlength{\arraycolsep}{1.5pt}
\NiceMatrixOptions%
{code-for-first-row=\scriptsize}
\[
\Gamma=\begin{pNiceMatrix}[first-row]
\rotatebox{45}{MemIn} & \rotatebox{45}{Relu} & \rotatebox{45}{Conv1} & \rotatebox{45}{Swish} & \rotatebox{45}{Conv2} & \rotatebox{45}{Add} & \rotatebox{45}{MemOut} & \\
\frac{BW}{2} & -R_{in_{RL}} & 0            & 0 				& 0 			& 0			 	& 0 \\
0            & R_{out_{RL}} & -R_{in_{C1}} & 0 				& 0 			& -R_{in1_{AD}}	& 0 \\
0            & 0            & R_{out_{C1}} & -R_{in_{SW}}	& 0 			& 0 			& 0 \\
0            & 0            & 0            & R_{out_{SW}} 	& -R_{in_{C2}}	& 0 			& 0 \\
0            & 0            & 0            & 0 	            & R_{out_{C2}} 	& -R_{in2_{AD}}	& 0 \\
0            & 0            & 0            & 0 				& 0          	& R_{out_{AD}}  & \frac{BW}{2}
\end{pNiceMatrix}
\] \\
\normalsize

The upper bi-diagonal structure of the $\Gamma$ matrix precludes modelling of branching behaviours, i.e. of graphs with nodes having multiple outgoing arcs as well as nodes accepting multiple incoming arcs. The suggested SDFG modification facilitates the creation of graphs with numerous incoming or outgoing arcs at nodes. The matrix below illustrates an example of the proposed SDFG. $BW$ is the available memory bandwidth to the device off-chip memory, which is given in the device specs that are provided as input to the system along with the high level network description file. The rest of the matrix elements indicates values that have emerged from the result of the matrix multiplication in the $\Gamma$ matrix.

The depth of each side of a branch is computed to incorporate some extra buffering for the streams that are combined at the merge points in order to ensure the right flow of data across the design's streams as well as to equalise the rates at the merge points.

\subsection{X3D layers as hardware building blocks} \label{hw_building_blocks}

The hardware building blocks are the major components utilised to construct the SDFG, which will be used subsequently to estimate the network's performance. The configuration of these blocks, in conjunction with the network's topology, is utilised to automatically generate and construct the design's synthesisable Vitis HLS code. The representation of the supported hardware building blocks comprises of the following: 
            \begin{enumerate}[i)]
                \item \textbf{DAG parameters}. A set of parameters that originated from the layer's settings as a DAG node. These settings are the layers' structural configuration that cannot be changed.
                \item \textbf{SDFG parameters}. An additional set of parameters that go along with the layer's hardware building block. These parameters have an impact on the layer's performance and are the ones that the optimization algorithm searches for during the design space exploration phase.
            \end{enumerate}
All of the layers share a common set of parameters describing their hardware building blocks. These parameters are the number of streams accepted at the layer's input and provided at the layer's output, namely $\mathbf{s_{in}}$ and $\mathbf{s_{out}}$, the consumption rate of the layer, defined as number of elements consumed per cycle $\mathbf{r_{in}}$, and the production rate of the layer, defined as number of elements produced per cycle $\mathbf{r_{out}}$. The detailed description of the parametrisation of all the layers is given below:
\begin{itemize}
        \item \textbf{3D Convolutional Layer}, $\mathbf{<{DAG_{params}}, {s_{in}, s_{out}, r_{in}, r_{out}, p_{mac}}>}$
            \begin{itemize}[label=$\circ$,leftmargin=+.2in]
                \item \textit{$p_{mac}$} is the number of multiply and accumulate (MAC) operations that take place inside the convolution in parallel.
            \end{itemize}
            The \textit{$s_{in}$},\textit{$s_{out}$}, and \textit{$p_{mac}$} are altered during the DSE and affect the final performance of the layer. Meanwhile the \textit{$r_{in}$} and \textit{$r_{out}$} depend on the \textit{$p_{mac}$} which means they are implicitly altered during the DSE as well. A mode detailed analysis of the convolution layer and its sub-modules is provided in fpgaConvNet \cite{Venieris2019}. 
        \item \textbf{3D Activation Layers, 3D Global Average Pooiling Layer}, $\mathbf{<{DAG_{params}}, {s_{in}, s_{out}, r_{in}, r_{out}}>}$ \\
            These layers' \textit{$r_{in}$} and \textit{$r_{out}$} can achieve consumption/production rates of 1 without constraints from previous layers or the memory, due to their element-wise functionality and the simplicity of their operations. The only exception here is the 3D Global Average Pooiling, in which $r_{out} = \frac{1}{D \times H \times W}$, where $D$ is the depth, $H$ is the height, and $W$ is the width dimension of the input feature map.
        \item \textbf{3D Element-wise Layers}, $\mathbf{<{DAG_{params}}, {s_{in1}, s_{in2}, s_{out}, r_{in1}, r_{in2}, r_{out}}>}$ \\
            This layer's \textit{$r_{in1}$}, \textit{$r_{in2}$} and \textit{$r_{out}$} can achieve consumption/production rates of 1 without constraints from previous layers or the memory, due to their element-wise functionality and the simplicity of their operations. It should be noted that in cases when the rates in either of the inputs are restricted owing to a lower production rate of a previous layer or due to memory constraints, the layer's input rates are equalised to the lower consumption rate among them.
\end{itemize}

\subsection{Streaming-Centric Optimizations}\label{x3d_optimizations}

During the development of the proposed design, the squeeze and excitation modules was found to hinder the throughput. Due to its averaging over the whole feature map, the 3D Global Average Pooling layer delays the design pipeline and demands increased branch stream buffering. The interdependence between these layers' output and the input of the ones that follow (the layer after GAP is always a point-wise convolution that requires all input channels to begin producing its output) and the shortage of on-chip memory due to the buffers' larger size limit performance. An approximation was introduced to solve this. GAP re-uses the results of its previous execution saved on the on-chip memory for each subsequent volume of data (i.e., every subsequent batch) and forwards them immediately to the next layer in the pipeline. Thus, the pipeline never stalls and buffering is greatly reduced. The GAP now calculates and saves the current batch's results on the on-chip memory without affecting the pipeline's other stages.

Because the authors of the original publication did not publish the model's performance in that specific benchmark, the model was first trained/fine-tuned before being evaluated on the UCF101 benchmark. Following the model's fine-tuning, the evaluation was carried out utilising the widely accepted strategy used on HAR benchmarks, which results in a total of 30 predictions that are averaged from a uniform sampling of 10 temporal clips for each video and spacial sampling of 3 crops in each frame. Furthermore, to be equivalent and directly comparable to previous studies, the impact of employing fixed-point precision has been investigated, by converting the model weights and all intermediate feature mappings from 32-bit floating point to 16-bit fixed-point arithmetic.

\begin{figure}[]
    \captionsetup{justification=centering}
    \centering
    \includegraphics[width=1\linewidth,keepaspectratio]{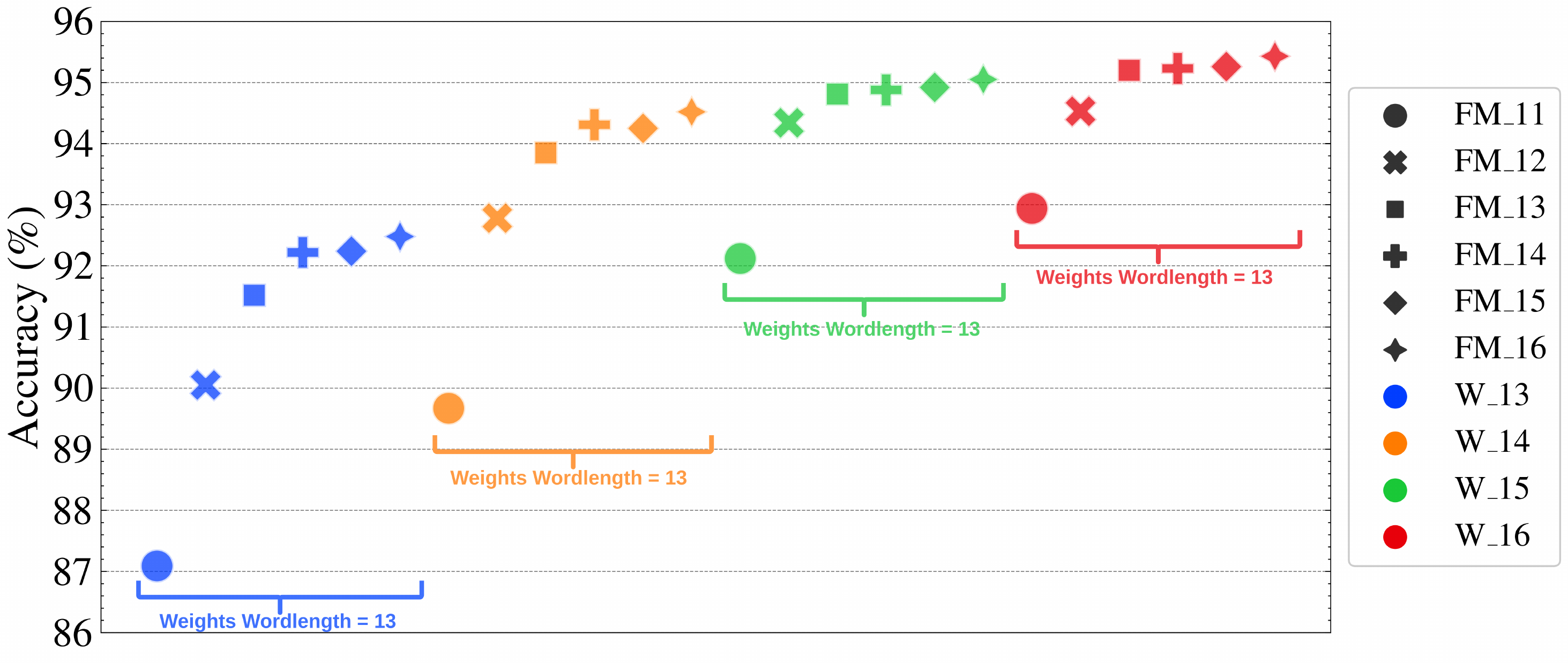}
    \caption{Quantization results on UCF101 over different word lengths. $W\_X$ denotes the word length of the fixed point representation of the weights, while $FM\_X$ for the feature maps}
    \label{fig:quantization_w_fm}
\end{figure}

Figure \ref{fig:quantization_w_fm} depicts the results of quantization of both the weights and feature maps over different combinations of word lengths. The experiments were conducted on the first split of UCF101 with a single clip evaluation (compared to 30 clip evaluation as described above). The accuracy achieved for this split with floating point feature maps and weights is 95.71\%. On a 16-bit fixed point precision for both the weights and feature maps we observe a drop in accuracy of 0.28\%. Moving to a 16-bit precision this drop increases to 0.66\%, while at 14-bits the accuracy drop is equal to 1.19\%. An interesting insight emerged from this evaluation is that the feature maps are less sensitive to reduction of the word length used to represent them compared to the weights of the model. For all the experiments we have explored and kept the best combination of the number of integer and fractional bits for each word length both for weights and feature maps. Finally we have evaluated the model on all three splits of UCF101 with a 30 clip evaluation using 16-bit fixed point for weights ($Q6.10$) and for feature maps ($Q7.9$). The accuracy drop was negligible dropping from 96.56\% to 96.43\%.

\section{Design Space Exploration}\label{dse}

The hardware mapping of the SDFG assumes a final streaming architecture to be inferred. Each design point in the design space has a specific combination of the involved layers' tunable parameters as they were described in section \ref{hw_building_blocks}. Essentially using a set of transformations to the SDFG the aforementioned parameters are being altered while the design space is being explored by simulated annealing, the heuristic optimization algorithm used in this study.

\subsection{X3D Model Partitioning}

There are primarily two techniques for designing CNN hardware architectures.
\begin{enumerate*}[(a)]
    \item Single computation engines rely on a more general and powerful processing unit that is time-shared across the network's layers, as well as a scheduler, to execute operations in the right sequence.
    \item Streaming architectures, such as the one presented, consist of a unique hardware block for each CNN layer that is optimised and exploits the parallelism of that layer independently of the others. In such designs, the more layers a CNN has or the larger the network's input gets, the greater the FPGA resource utilisation becomes, which may limit the achieved parallelism of each layer in order to fit all layers into a single design.
\end{enumerate*}

Splitting the network execution into smaller partitions can solve this kind of problem while at the same time exploiting the reconfiguration capabilities of the FPGA. Furthermore, by producing a unique architecture and delivering a separate bitstream for each partition, it allows the design of more finely tuned architectures that are better tailored to each layer's characteristics. This approach also drastically reduces off-chip memory access to only the design's input and output streams, allowing the on-chip memory to be used for data reuse. Following this strategy, the cost of reconfiguration must be considered whenever a new partition is loaded onto the board, although this cost may be mitigated by increasing the amount of data each partition have to process (i.e. by increasing the batch size).

\begin{figure}[]
    \captionsetup{justification=centering,margin=1.25cm}
    \centering
    \begin{subfigure}[c]{.3\linewidth}
        \centering\includegraphics[width=0.9\linewidth,height=6cm,keepaspectratio]{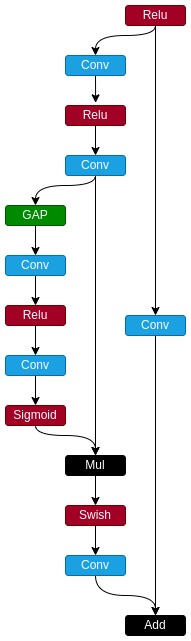}
        \caption{}
    \end{subfigure}
    \begin{subfigure}[c]{.3\linewidth}
        \centering\includegraphics[width=0.9\linewidth,height=6cm,keepaspectratio]{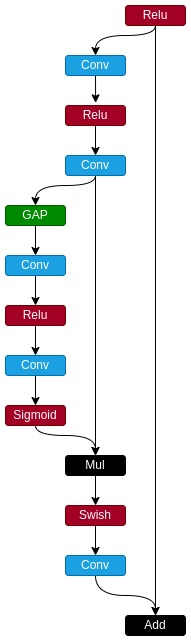}
        \caption{}
    \end{subfigure}
    \begin{subfigure}[c]{.3\linewidth}
        \centering\includegraphics[width=0.55\linewidth,keepaspectratio]{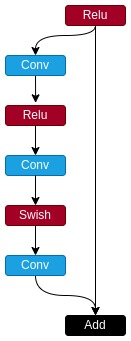}
        \caption{}
    \end{subfigure}
    
    \caption{X3D model main partitions types: (a) Type 1 (b) Type 2 (c) Type 3}
    \label{fig:x3d_layer_types}
\end{figure}

The partitions in this work are determined based on three primary layer types identified in the X3D model. Figure \ref{fig:x3d_layer_types} depicts the structure of the suggested layer types.

\setlength{\belowcaptionskip}{-7pt}
\setlength{\belowcaptionskip}{-15pt}

The X3D model comprise of 26 such layers, 4 of which are of layer type 1, 11 of which are of layer type 2, and 11 of which are of layer type 3. Each layer type, as can be observed from their structure, displays branching behaviour, with type 1 and 2 exhibiting double branching owing to the squeeze and excitation module \cite{Hu2020}.

\subsection{DSE Within X3D Partitions}

Two factors determine the parallelism of each X3D partition layer's tunable parameters. The first is the number of parallel coarse operations in each layer, which depends on the input feature map's channels. The primary operations of each layer can be executed in parallel by deploying several instances of the processing block up to the number of channels because the operations in each channel dimension are independent. Both input channels and output filters can use coarse-level parallelism in 3D convolutional layers. The $s_{in}$ and $s_{out}$ parameters of the hardware building block configuration are updated and searched for optimal values throughout the DSE to achieve this parallelism, affecting the stream matrix $\textbf{S}$, the topology matrix $\Gamma$, and therefore the design performance.

The second parallelism factor affects the 3D convolutional layers and determines the dot product operation's parallelism during the kernel's convolution with an input volume. This parallelism specifies how many multipliers will be parallel for multiplications and how many levels on the adder tree for additions. A completely unrolled configuration uses N multipliers and N-1 adders, producing 1 dot product per cycle. Limiting the setup to a single multiplier and adder produces 1/N dot products per cycle, where N is the shape of the flattened input and kernel. As can be evident, a trade-off exists between throughput and resource utilisation. The DSE optimises the $p_{mac}$ parameters of the 3D convolutional layer hardware building block configuration to achieve this parallelism, altering the rate matrix $\textbf{R}$, the topology matrix $\bm{\Gamma}$, and defining the design performance.

\subsection{Performance Modeling}

To describe the performance of a given design based on the topology matrix $\Gamma$, an additional matrix reflecting the workload of each layer must be included. Because the topology matrix provides the throughput of each layer at its input in consumptions/cycle and output in productions/cycle, constructing a matrix with the total workload of each layer, i.e. the total number of elements to be consumed and produced, will allow the generation of a new matrix that provides the number of cycles each layer requires to consume its workload. More specifically a workload matrix $\textbf{W}$ has the same structure as the topology matrix $\Gamma$ in terms of nodes and arcs of the SDFG. The values of this matrix indicate the total number of elements to be consumed at the input of each layer as well as the total number of elements to be produced at the output of each layer. By element-wise dividing the $\textbf{W}$ matrix with the $\Gamma$ matrix, the final $II$ matrix is being calculated as shown below:
\begin{equation}\label{ii_matrix}
    II = W / \Gamma
\end{equation}
The $II$ is the initiation interval matrix, and its entries represent the total number of cycles required by each layer to consume its workload completely. The maximum value of the $II$ matrix, denoted by $II_{max}$ determines the initiation interval of the whole SDFG. The total execution time of a partition with batch size B is given by the following equation:
\begin{equation}\label{exec_time}
    \operatorname{t(B,\Gamma)}=\frac{1}{\textrm{clock rate}}\cdot (D + II_{max}\cdot (B-1))
\end{equation}
where $D$ is the total number of cycles needed to fill the pipeline depth of the whole design, and its calculated by adding the individual depths of each layer and the extra depth added due to the extra buffering to deal with the branches in the design.

In order to capture the model's overall execution time, the execution times of each individual partition are summed up with the addition of the total reconfiguration time, as shown below:
\begin{equation}\label{exec_time_total}
    \operatorname{t_{total}(B,\Gamma)}=\sum_{n=0}^{N_p}t_n(B,\Gamma_i) + (N_p-1)\cdot t_{reconfig}
\end{equation}
where $N_p$ is the total number of the partitions of the model, and $t_{reconfig}$ is the reconfiguration time needed before loading each partition to the FPGA. As can be noticed from Eq. \ref{exec_time_total}, the extra overhead caused by the device reconfiguration is proportional to the number of partitions of the final solution and is independent of the batch size. By increasing the number of batches processed by the model, the first term dominates the execution time and the cost of reconfiguration is amortised.

Finally the overall throughput of the proposed architecture is inferred by dividing the total workload of the model in GOPs (Giga Operations) times the batch size, with the total execution time:
\begin{equation} \label{tota_throughput}
    \operatorname{Throughput(B)}=\frac{Workload_{model}*B}{t_{total}(B,\Gamma)}
\end{equation}

The design space exploration on each partition is described as an optimization problem with the following objective: $max(t(B,\Gamma)), s.t. rsc(\Gamma) \leq rsc_{avail}$. The simulated annealing \cite{1993ModernProblems} algorithm, a heuristic approach used to tackle this optimization problem, attempts to optimise the design's throughput while ensuring that FPGA resource use does not exceed the available resources (as these provided in the device specifications)

\section{Evaluation} \label{evaluation}

For evaluation purposes, the ZCU102 has been used as the target platform. In all of the experiments the target frequency was set to 142 MHz. Vitis HLS and Vivado Design Suite (v21.2) were used, while the reported resource results are after place and route. In all the designs the arithmetic precision used was 16-bit fixed point arithmetic as describen in Section \ref{x3d_optimizations}. The UCF-101 \cite{Soomro2012UCF101:Wild} HAR benchmark was used to assess the correctness of the X3D model and position it in relation to prior studies. Along with the baseline architecture, a second one is proposed that uses statistics from previous batches of GAP layers.

%

\subsection{Model Prediction Accuracy Evaluation}

\begin{figure}[]
    \captionsetup{justification=centering,margin=1.25cm}
    \centering
    \includegraphics[width=0.85\linewidth,keepaspectratio]{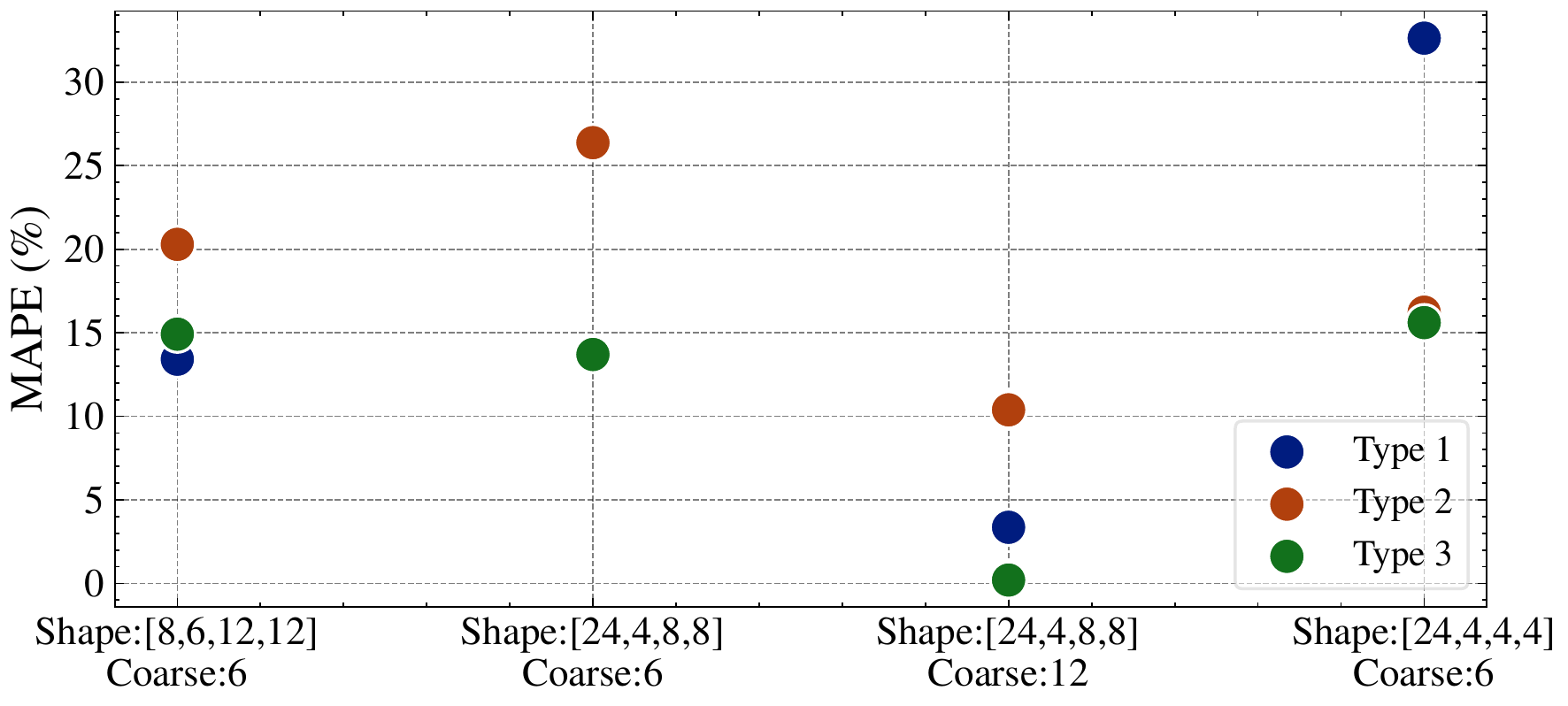}
    \caption{Latency (ms) MAPE between modeling performance and co-simulation}
    \label{fig:modeling_errors}
\end{figure}

To ensure that the performance of X3D predicted by the modelling is representative of the actual performance provided by co-simulation results, a series of experiments has been conducted for each supported layer type with different configurations for the layer input and output shapes, as well as the coarse in and coarse out factors. The metric of Mean Absolute Percentage Error (MAPE) was used to measure the error on the predicted vs co-simulation latency for all the experiments. As can be seen from Figure \ref{fig:modeling_errors}, the errors vary depending on the coarse in and coarse out factors; the bigger the coarse factors get the smaller the error becomes. On average, layer type 1 has an MAPE of 11.92\%, layer type 2 has 17.32\%, and layer type 3 has 5.03\%.

\subsection{Performance Evaluation}

\begin{figure}[h]
    \captionsetup{justification=centering,margin=1.0cm}
    \centering
    \includegraphics[width=1.0\linewidth,keepaspectratio]{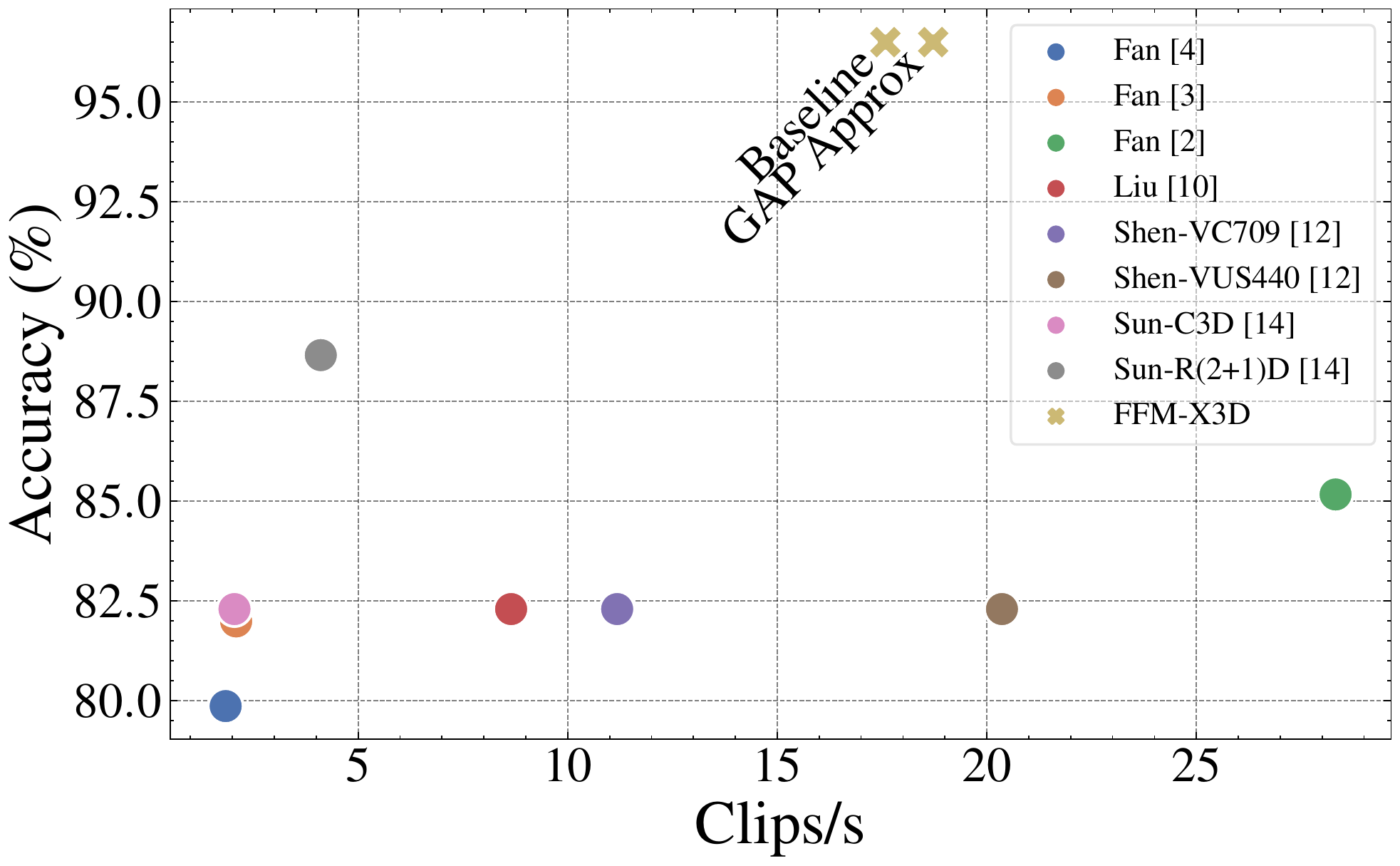}
    \caption{Clips per second, positioning of FFM-X3D within the test of the existing works}
    \label{fig:clips_per_sec_batch_100}
\end{figure}

To the best of our knowledge, no previous works have targeted the acceleration of X3D onto FPGA as of the date of authoring this study. To position within the previous works, and because most current works focus on the C3D model targeting different FPGA platforms, the assessment is being undertaken on a number of different metrics: 
    \textbf{Clips per second (Clips/s)}. This metric is the total number of clips processed over a second. The values on this metric on Table \ref{performance_comparison} are measured and repodted for a batch size of 100 as denoted in the Table's footnotes. The positioning of the FFM-X3D related to the existing works can be seen in Figure \ref{fig:clips_per_sec_batch_100}. The proposed architecture is positioned in the pareto front of accuracy over clips/s compared to the existing works in the field.
    \textbf{Throughput (GOp/s)}. To be fully aligned with the results of the existing works, GOp/s are accounted as the MAC (multiply and accumulate) operations per second. For batch size 100, FFM-X3D achieves better throughput compared to \cite{Fan2017,Fan2018,Sun20203DPruning} while at the same time achieves the better accuracy among all the existing works.
    \textbf{DSP efficiency (GOp/s/DSP)}. This metric is the normalized throughput over the DSPs used. In this work, the DPSs used are averaged over all the 26 pre-defined partitions. A positioning of FFM-X3D in relation to the existing works for a favorable batch size of 100, shows that FFM-X3D does provide comparable DSP efficiency to most of the works.

Given that the proposed design focuses on throughput, the related works are also evaluated based on this metric. A batch size 100 was chosen for FFM-X3D architecture to amortise the reconfiguration cost. These batch size values can be used in systems that analyse multiple videos in parallel, or in evaluation processes that use multiple clips from a single video. In comparison to M. Sun et. al. \cite{Sun20203DPruning}, the sole effort targeting the same FPGA platform, FFM-X3D outperforms its C3D and R(2+1)D counterparts by $1.07\times$ and $1.52\times$, respectively, while latency is $4.54\times$ and $9.11\times$ lower. Table \ref{performance_comparison} summarizes the positioning of FFM-X3D in relation to the rest of the existing works.

\setlength{\belowcaptionskip}{0pt}
\begin{strip}
\centering
\captionof{table}{Positioning of FFM-X3D in the space of 3D CNN HAR models} \label{performance_comparison}
\begin{adjustbox}{width=1.\textwidth}
\begin{threeparttable}
\begin{tabular}{cc|c|c|c|cc|cc|cc}
\hline
& H. Fan \cite{Fan2017} & H. Fan \cite{Fan2018} & H. Fan \cite{Fan2019} & Z. Liu \cite{Liu2019AFpgas} & \multicolumn{2}{c|}{J. Shen \cite{Shen2018}\tnote{$\ddagger$} } & \multicolumn{2}{c|}{M. Sun \cite{Sun20203DPruning}} & \multicolumn{2}{c}{FFM-X3D (Ours)} \\
\cline{6-11}
& & & & & VC709 & VUS440 & C3D & R(2+1)D-18 & Baseline & GAP-approx \\
\hline  \hline
FPGA                & ZC706 & ZC706 & Intel SX660 & VC709 & VC709 & VUS440 & ZCU102 & ZCU102 & ZCU102 & ZCU102 \\
Model               & C3D & C3D & E3D & C3D & C3D & C3D & C3D & R(2+1)D-18 & X3D-M & X3D-M \\
GFLOPS \tnote{$\ast$}    & 38.61 & 38.61 & 6.1 & 38.61 & - & - & 38.61 & 26.14 & 6.4 & 6.4 \\
\rowcolor{lightgray} Accuracy            & 79.87 \% & 81.99 \% & 85.17 \% & 82.3 \% & 82.3 \% & 82.3 \% & 82.3 \% & 88.66 \% & 96.5 \% & 96.5 \% \\
\rowcolor{lightgray} Clips/s \tnote{$\dagger$}              & 1.84 & 2.09 & 28.32 & 8.65 & 11.18 & 20.36 & 2.05 & 4.11 & 17.58 & 18.72 \\
GOp/s \tnote{$\dagger$}              & 70.41 & 80.12 & 172.8 & 330.74 & 427.29 & 778 & 78.44 & 111.71 & 112.41 & 119.83 \\
GOp/s/DSP  \tnote{$\dagger$}         & 0.087 & 0.103 & 0.109 & 0.092 & 0.281 & 0.511 & 0.065 & 0.092 & 0.052 & 0.055 \\
Clock (MHz)         & 172 & 200 & 150 & 120 & 150 & 200 & 150 & 150 & 142 & 142 \\
Precision           & 16-bit fixed & BFP & 32-bit float & 16-bit fixed & 16-bit fixed & 16-bit fixed & 16-bit fixed & 16-bit fixed & 16-bit fixed & 16-bit fixed \\
DSP \%            & 90 \% & 86.6 \% & 93.3 \% & 99.8 \% & 42 \% & 53 \% & 48 \% & 48 \% & 86 \% & 85 \% \\
\hline
\end{tabular}
\begin{tablenotes}
    \item[$\ast$] GFLOPS are calculated as MACs.
    \item[$\dagger$] Favorable batch size 100.
    \item[$\ddagger$] The C3D model used is different/smaller version from the original one \cite{Ji2013}.
\end{tablenotes}
\end{threeparttable}
\end{adjustbox}
\end{strip}

Table \ref{performance_comparison_gpu} shows a performance comparison of the final FFM-X3D design  on a ZCU102 FPGA platform versus a state of the art server graded GPU platform. A GeForce RTX 3090 with 10496 CUDA cores is the target GPU utilizing CUDA and CuDNN libraries to optimise its solution. The proposed solution has 4.5x longer latency, but with a significant reduced clock rate.

\begin{table}[]
\centering
\caption{Performance comparison versus GPU}\label{performance_comparison_gpu}
\resizebox{0.8\columnwidth}{!}{%
\begin{tabular}{ccc}
\hline
& GPU & FFM-X3D (Ours) \\
\hline
Platform                & GeForce RTX 3090 & ZCU102 \\
Frequency               & 1.7 GHz & 142 MHz \\
Power (W)               & 257.9 & 26 \\
Batch Size              & 1 & 1 \\
Precision               & 32-bit float & 16-bit fixed \\
Clips/s/Watt (cps/W) & 0.254 & 0.0177 \\
\hline
\end{tabular}
}
\end{table}

\section{Conclusion} \label{conclusion}

This paper introduces FFM-X3D, a hardware architecture for X3D, a 3D CNN model for HAR that provides state of the art performance. The proposed methodology follows the fpgaConvNet paradigm making use of the SDF theory to describe and map X3D into hardware designs. The revised $\Gamma$ matrix construction allows the SDF modelling approach to deal with CNN topologies that contain branches, something that is being taken into account throughout the design space exploration . Experiment results reveal that the FFM-X3D provides competitive throughput and DSP efficiency while attaining up to $16.63$\% better prediction accuracy on the selected HAR benchmark. Future work might involve expanding the design space with extra SDFG transformations like automated partition generation, as well as generalising to enable the support of a broader collection of 3D CNN models, expanding to other video or volume related tasks beyond the HAR.

\section*{Acknowledgment}
For the purpose of open access, the authors have applied a Creative Commons Attribution (CC BY) license to any Accepted Manuscript version arising.

\bibliographystyle{acm}
\bibliography{references}

\end{document}